\documentclass[10pt,twocolumn,letterpaper]{article}

\usepackage{wacv}
\usepackage{times}
\usepackage{epsfig}
\usepackage{graphicx}
\usepackage{amsmath}
\usepackage{amssymb}
\usepackage{mathtools}
\usepackage{diagbox}
\usepackage{booktabs,siunitx}
\usepackage{adjustbox}

\wacvfinalcopy 

\def\wacvPaperID{1298} 

\ifwacvfinal\fi
\begin{document}

\title{CineFilter: Unsupervised Filtering for Real Time Autonomous Camera Systems}



\author{ 
\emph{Sudheer Achary$^{\dag}$},  \emph{K L Bhanu Moorthy$^{\dag}$}, 
\emph{Ashar Javed$^{\ddag}$}, \emph{Nikita Shravan$^{\dag}$}, \\ 
\emph{Vineet Gandhi$^{\dag}$}, \emph{Anoop Namboodiri$^{\dag}$} \\ \\
{ $^{\dag}$CVIT, KCIS, IIIT Hyderabad, $^{\ddag}$ Carnegie Mellon University }\\
}


\maketitle
\ifwacvfinal\thispagestyle{empty}\fi

\begin{abstract}
Autonomous camera systems are often subjected to an optimization/filtering operation to smoothen and stabilize the rough trajectory estimates. Most common filtering techniques do reduce the irregularities in data; however, they fail to mimic the behavior of a human cameraman. Global filtering methods modeling human camera operators have been successful; however, they are limited to offline settings. In this paper, we propose two online filtering methods called Cinefilters, which produce smooth camera trajectories that are motivated by cinematographic principles. The first filter (\emph{CineConvex}) uses a sliding window-based convex optimization formulation, and the second (\emph{CineCNN}) is a CNN based encoder-decoder model. We evaluate the proposed filters in two different settings, namely a basketball dataset and a stage performance dataset. Our models outperform previous methods and baselines on both quantitative and qualitative metrics. The \emph{CineConvex} and \emph{CineCNN} filters operate at about 250fps and 1000fps, respectively, with a minor latency (half a second), making them apt for a variety of real-time applications.

\end{abstract}

\section{Introduction}

Autonomous camera systems that track a person, object, or action of interest often employ a filtering operation on top of raw per frame estimations (e.g., a virtual camera following an instructor by cropping a window from a static camera around him every frame). Trajectory optimization is also useful in video stabilization methods, where the video is re-rendered with the stabilized trajectory. The main goal of trajectory optimization/filtering models is to transform the original trajectory to one that mimics the behavior of a skilled cameraman, as closely as possible. The cinematographic literature~\cite{thompson2013grammar} suggests that a good pan/tilt shot should comprise of three components: a static period of the camera at the beginning, a smooth camera movement, which ``leads" to the movement of the subject and a static period of the camera at the end. The prior art has shown~\cite{grundmann2011auto, gandhi2014multi, tang2019joint} that such a behavior can be modeled as an optimization problem leading to trajectories with piece-wise constant, linear, and parabolic segments. 

However, most of these methods~\cite{grundmann2011auto, gandhi2014multi, tang2019joint} are offline and require the entire trajectory during the stabilization process. Such offline filtering/optimization methods cannot be adopted for real-time applications such as live broadcasts (e.g., sports, lectures, live performances). Chen~\cite{chen2016learning} made an attempt to tackle this problem in real-time by learning a regression model using a specific ground truth signal generated by a human operator for a basketball game dataset. The learned model however, is inherently tied to the behavior specific to the particular basketball setting and its ground truth. Such a supervised approach is not applicable to arbitrary trajectory optimization. 


\begin{figure*}
    \centering
    \begin{tabular}[b]{c c c}
        \includegraphics[width=0.32\linewidth]{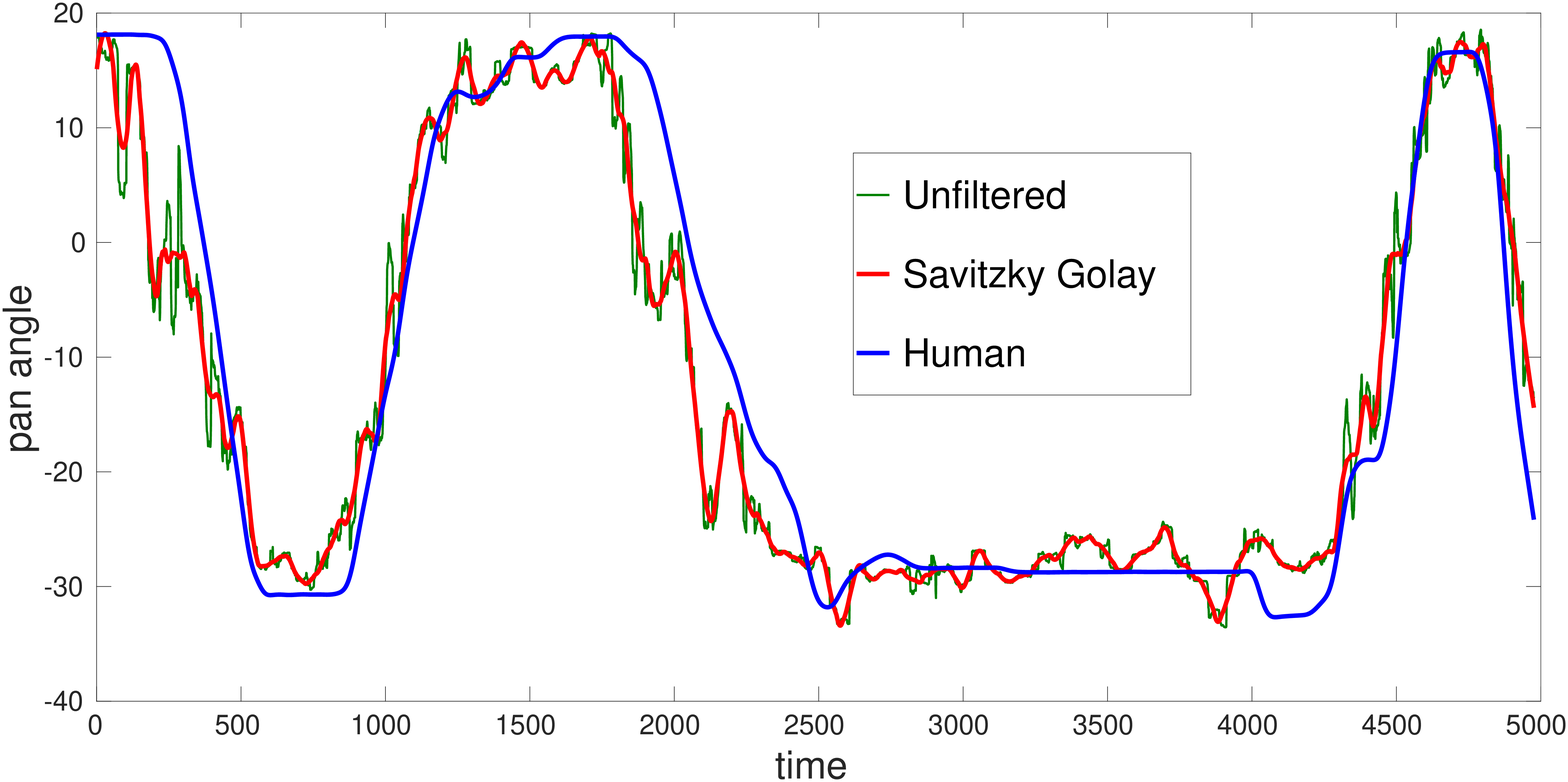}& \hspace{-1em}
        \includegraphics[width=0.32\linewidth]{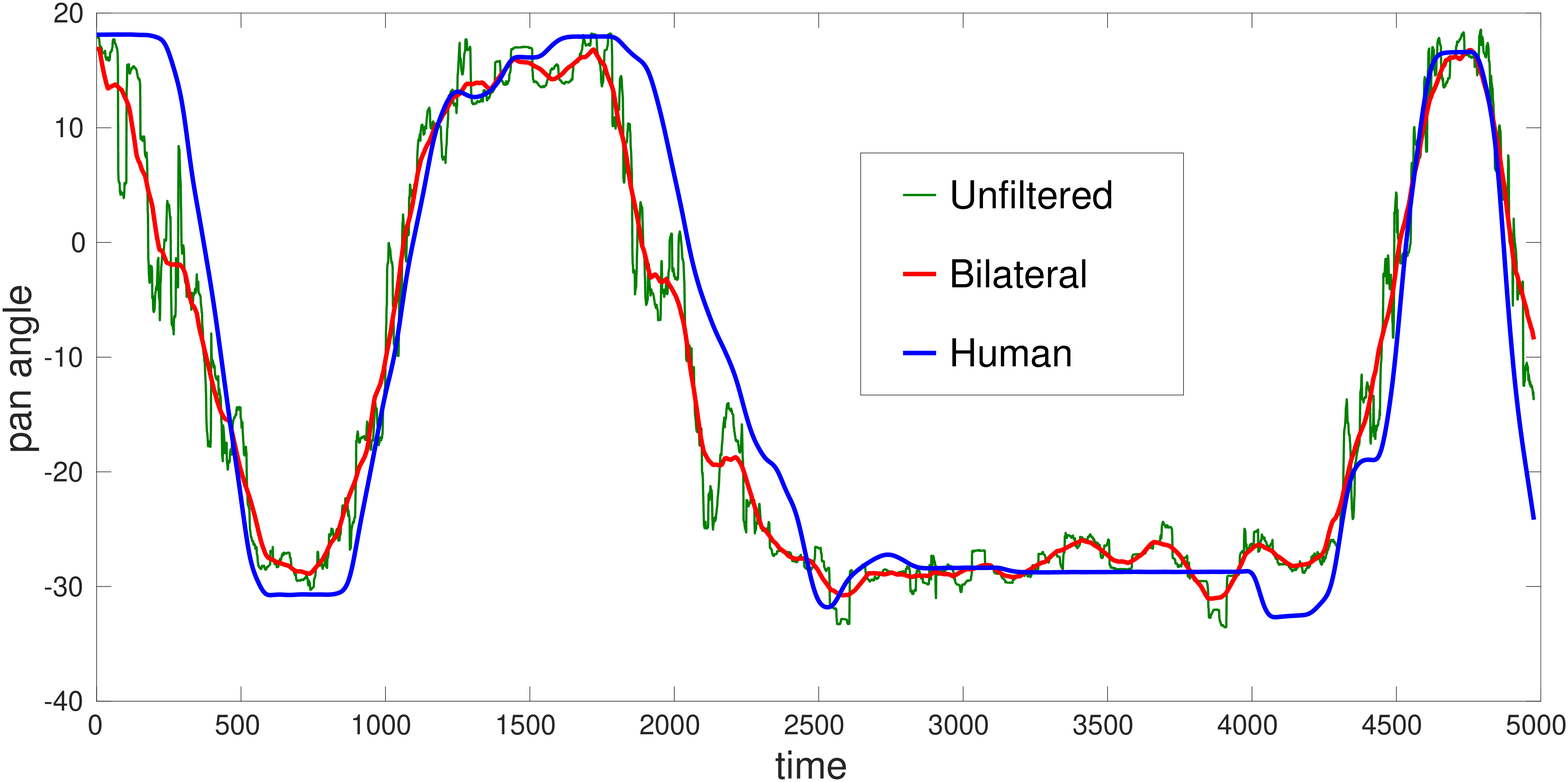}& \hspace{-1em}
        \includegraphics[width=0.32\linewidth]{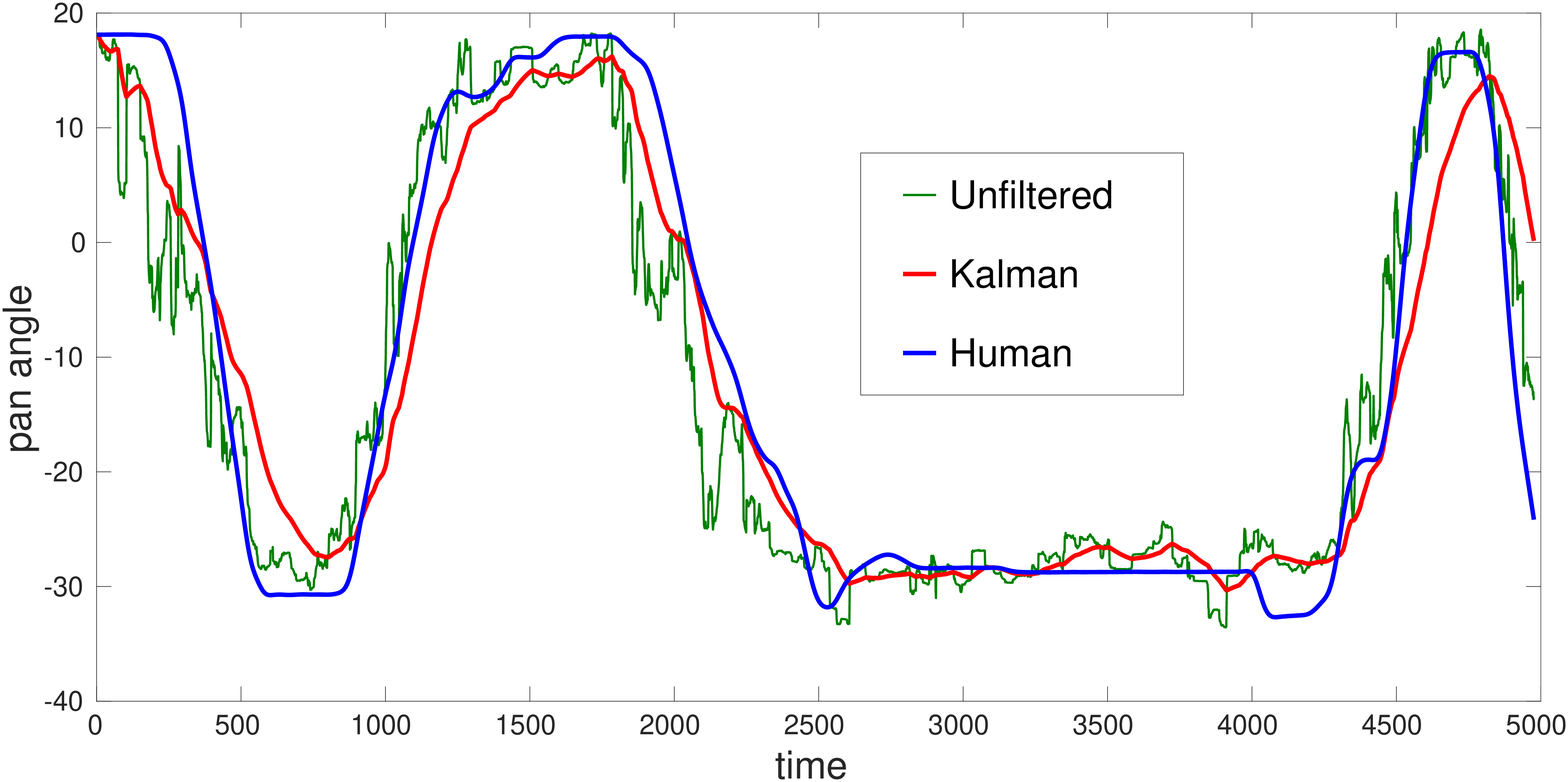} \\
        \includegraphics[width=0.32\linewidth]{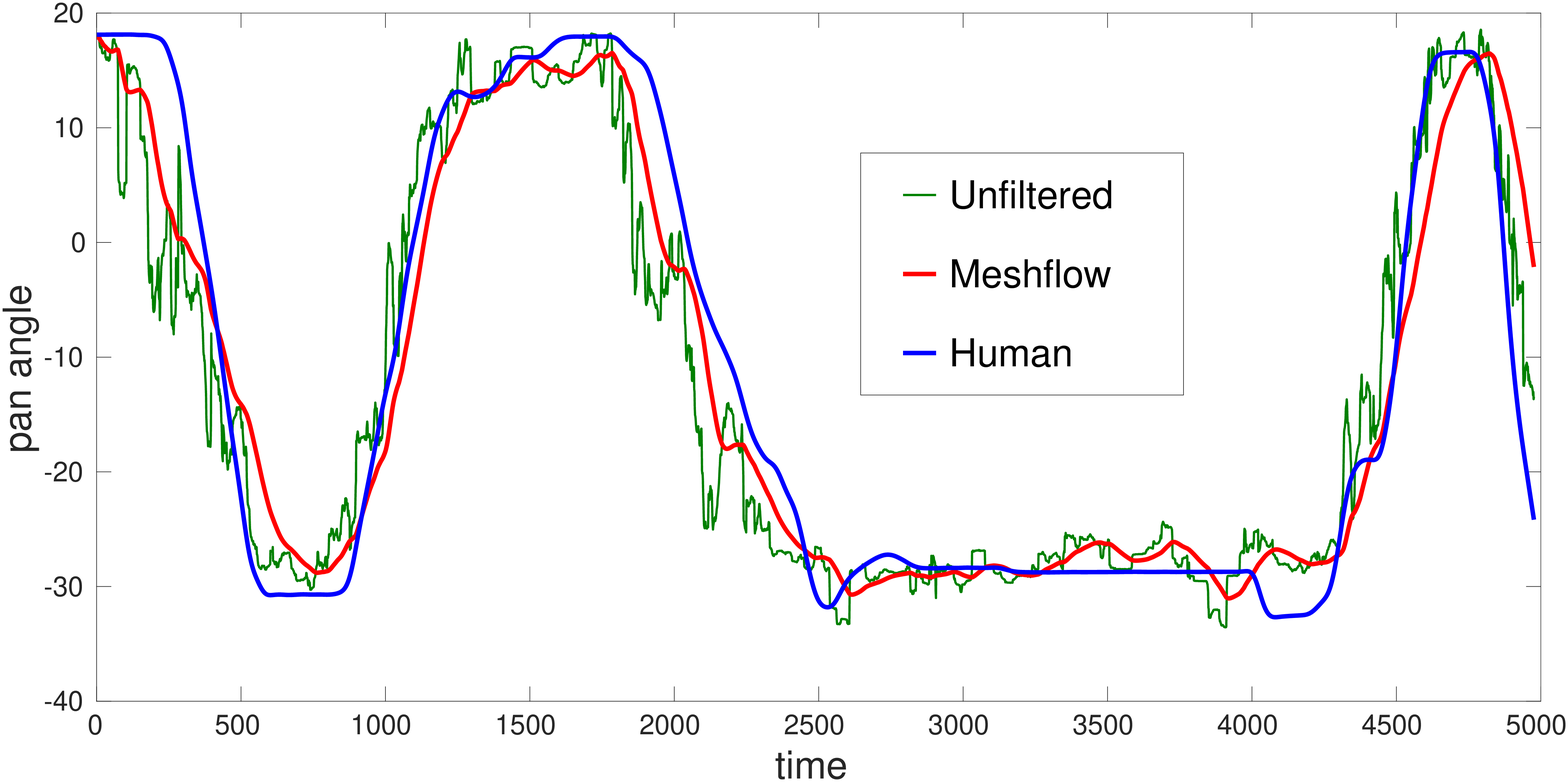}& \hspace{-1em}
        \includegraphics[width=0.32\linewidth]{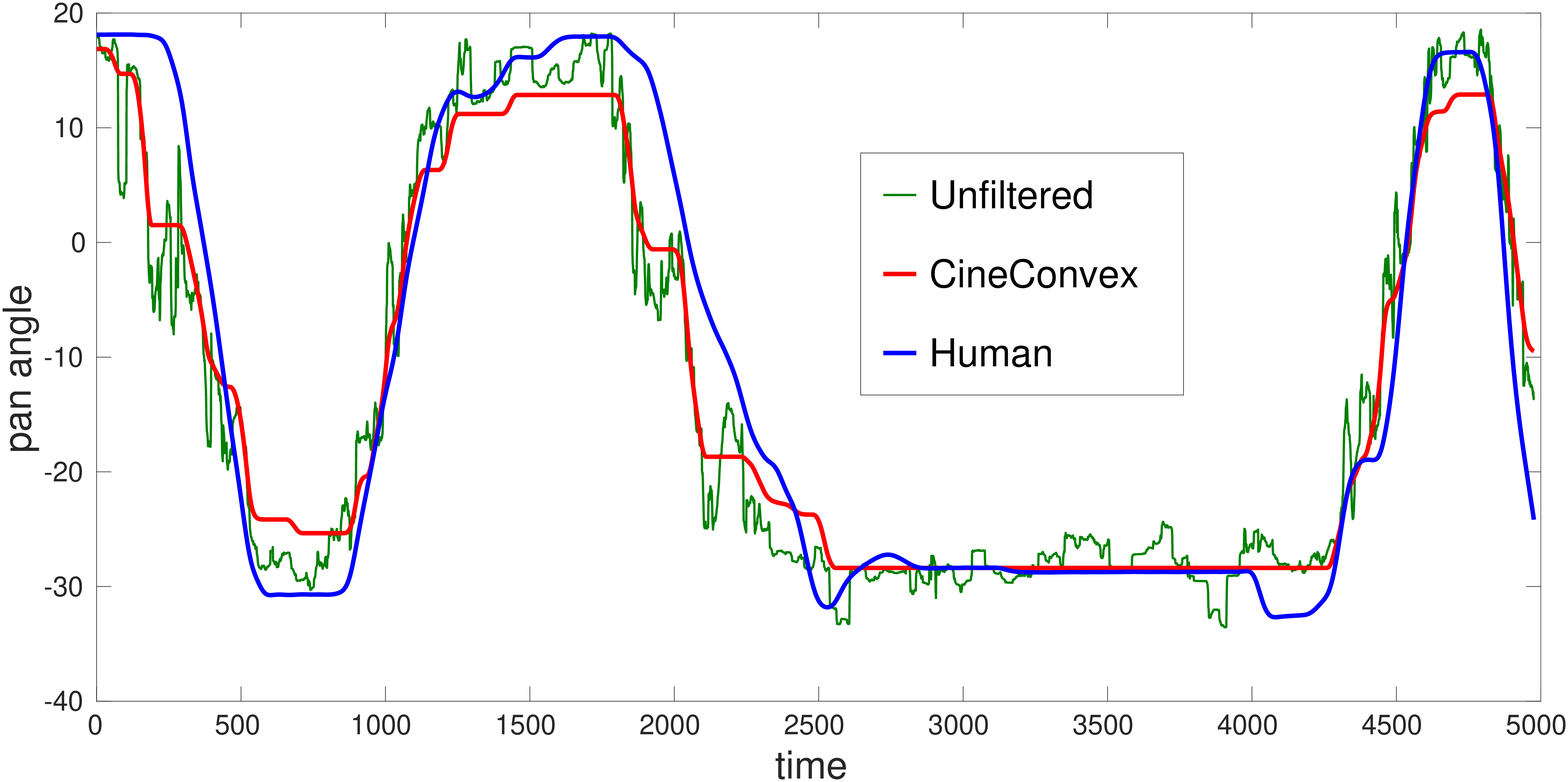}& \hspace{-1em}
        \includegraphics[width=0.32\linewidth]{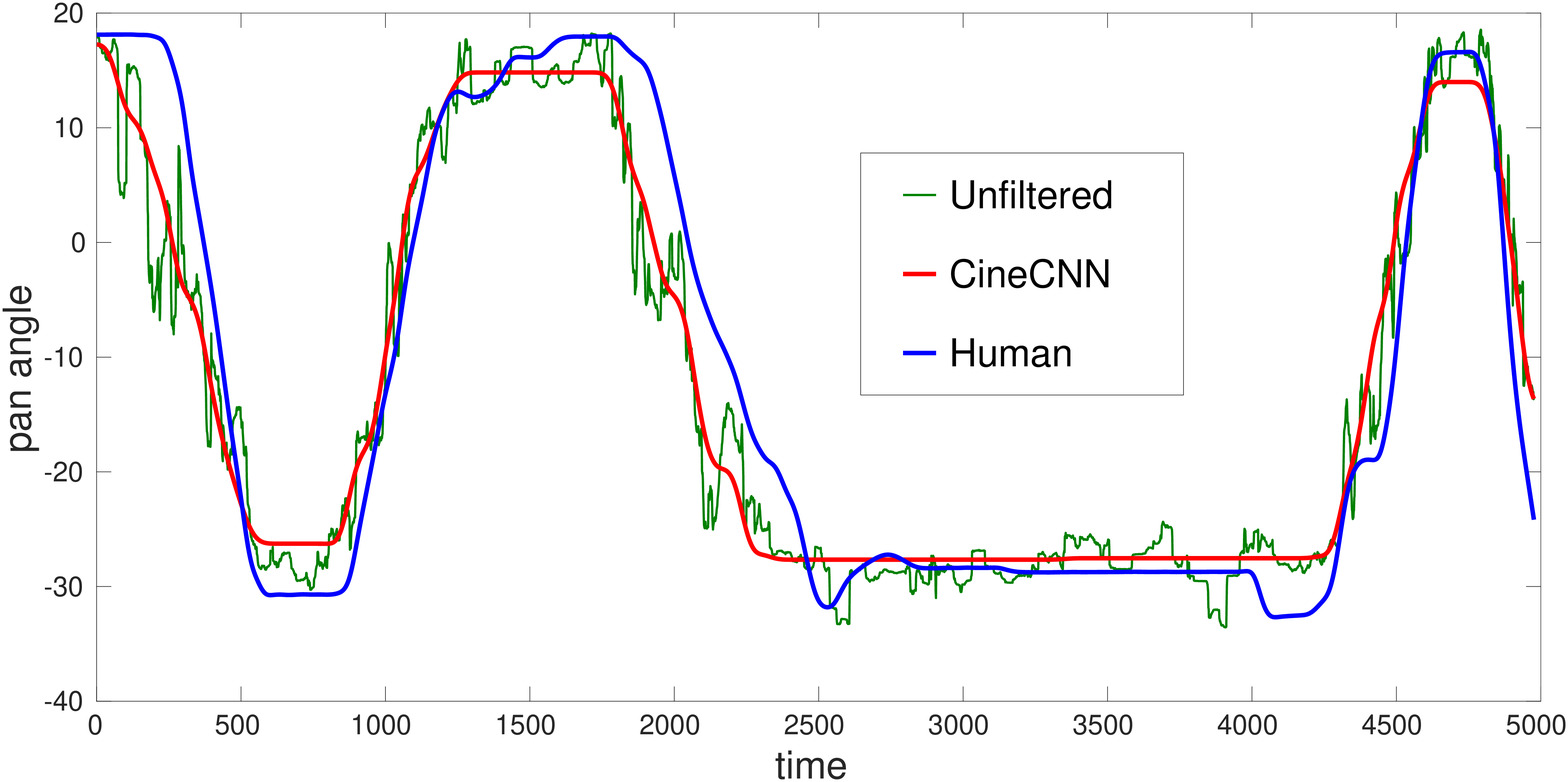} \\
    \end{tabular}
    \caption{The pan angle of a human cameraman compared to those generated by filtering per frame pan angle predictions using different algorithms over a basketball video sequence.The plots show the filtered outputs corresponding to several baseline algorithms and our proposed approach. The proposed CineFilters mimic professional cameraman-like behaviour and results in piece-wise static, linear and parabolic segments. Other filters fail to give perfect static segments, have sharp corners and sudden direction changes. Although they smooth the trajectories well, they are not appropriate for applications in autonomous camera systems.
    }
    \label{fig:teaser}
\end{figure*}

Real-time prediction of smooth camera trajectories is an ill-posed problem since, at any given instance, the predicted trajectory can be smoothed in multiple ways depending on the future input sequences. Thus, using direct optimization on trajectories like previous offline works in a piece-wise manner (sliding window) for the online prediction can make the system very brittle. In this paper, we study the effect of allowing a little peek into the future and applying causal constraints on pre-stabilized trajectories. We show that the proposed sliding window formulation of \emph{CineConvex} filter can closely mimic an offline global optimization with a latency of just half a second. We further investigate parametric models trained with data that can learn the multimodal distribution of a smooth camera trajectory in an online setting. We hypothesize that such predictive models can reduce the dependence on the latency, especially in high noise scenarios. Furthermore, the prediction itself can be done at much faster rates.

To this end, we propose \emph{CineCNN}, a novel CNN based filtering approach that can learn from patterns existing across multiple samples in the dataset. A key issue when training such models is the availability of large labeled datasets. However, our training is unsupervised and uses a set of generic objective functions on top of total variation denoising. Thus, we do not require any ground truth labels for training, and \emph{CineCNN} is independent of the idiosyncrasies of human labels from specific settings. Moreover, our objective functions also encourage cinematographically motivated behaviors like the prediction of true static segments without any residual motions and avoidance of sudden jerks, which is something missing from the past works. A motivating illustration is shown in Figure~\ref{fig:teaser}, where we compare our approach with filtering mechanisms prevalent in previous works. In summary, we make the following contributions:
\begin{enumerate}
    \item We propose \emph{CineConvex} formulation to adapt offline trajectory optimization~\cite{grundmann2011auto, gandhi2014multi, tang2019joint} into an online one. Filtering is posed as a convex optimization in sliding window fashion with minor latency and constraints on the optimized trajectory from the previous window. 
    \item We propose \emph{CineCNN} for learning a camera trajectory smoothing function. A total variation denoised input is sent to a convolutional neural network to convert into trajectories which mimic cameraman behaviour. The predictive model offers various advantages, especially in terms of computation load and robustness to noise.
    \item Our models can predict smooth trajectories in real-time (1000 FPS for the \emph{CineCNN} and 250 FPS for \emph{CineConvex} on an Intel Xeon CPU). \emph{CineCNN} has a low model complexity (400KB in size), which enables easy deployment in live settings.
    \item Both our approaches work in an unsupervised way and do not need any expensive ground truth labels as compared to previous approaches~\cite{chen2016learning}. This makes our model-training independent of the setting in which the label was obtained, thus making it easier to extend to new settings.
    \item Our quantitative and qualitative evaluations show that our approach can mimic professional cameraman behavior and outperforms the baselines and the prior art.
\end{enumerate}

\section{Related Work}
Research in autonomous camera systems dates back to more than two decades. One of the earliest systems was proposed by Pinhanez and Bobick~\cite{pinhanez1995intelligent}, which aimed at automated camera framing in cooking shows. Their system was based on two types of cameras, a spotting camera that watched the entire area of interest and a robotic tracking camera that followed the verbal instructions from a director to frame the desired targets automatically. The Autoauditorium~\cite{bianchi2004automatic} system extended this idea for lecture videos (a single presenter in front of a screen). The robotic camera in the Autoauditorium system crudely followed the presenter, whose position was estimated each frame using background subtraction on the spotting camera. 

A series of works then followed, and we review the computational models for camera movement, proposed in these approaches. Yokoi {\it et al.}~\cite{yokoi2005virtual} present a method for automated editing of lecture videos. Their work replaces the robotic camera by a virtual camera, i.e., a cropping window moving inside a high-resolution video. They use temporal frame differencing to detect the Region of Interest (RoI) around the presenter and use bilateral filtering to remove the jittery motion introduced by per-frame RoI estimations.  A similar digital tracking approach was employed in Microsoft's icam2 system~\cite{zhang2008automated}. The contemporary work by~\cite{sun2005region} also tracks cropped RoI from a panoramic video; however, it employs a Kalman filter for removing the jitter. Such filtering approaches successfully reduce jitter; however, they are not cinematically inspired, and they lead to unmotivated camera movements and fail to keep the camera static. Heuristics have been applied to tackle some of these challenges. For instance, \cite{sun2005region} keeps RoI unchanged if the Kalman filter predictions of new positions are within a specified distance of the registered position, and the new estimated velocity is below a threshold. However, such heuristics are not applicable in generalized scenarios. 

The virtual camerawork has been investigated for the autonomous broadcast of sports. Unlike the classroom environment, where framing a single person is relatively simple, the sports videos have multiple players and fast-moving objects. Diago {\it et al.}~\cite{daigo2004automatic} propose an offline system that uses the audience face direction to build an automatic pan control system (pan angle of the broadcasting camera). Ariki {\it et al.}~\cite{ariki2006automatic} proposed a heuristic-based editing strategy. However, these approaches focus on the per-frame pan angle estimation and skip details on the camera motion models or smoothing algorithms. Chen {\it et al.}~\cite{chen2010personalized} uses Gaussian Markov Random Fields (MRF) to obtain smooth virtual camera movements. They induce smoothness by inducing inter-frame smoothness priors. Recent work \cite{rachavarapu2018watch} has shown that while inter-frame smoothness priors do induce smoothness, it fails to give aesthetically pleasing camera behavior. They augment it with an additional post-processing optimization to render professional-looking camera trajectories. 

The virtual videography~\cite{heck2007virtual} system was one of the earliest works to model camera movement inspired by filmmaking literature~\cite{cantine1995shot, thompson2013grammar}. They define that a good tracking shot should consist only of smooth motions in a single direction that accelerate and decelerate gradually to avoid jarring the viewer while maintaining the correct apparent motion of the subject. They define a customized parametric function for the motion model and solve for parameters for each moving shot individually. Liu {\it et al.}~\cite{liu2006video} demonstrate the applicability of such a motion model in the application of automatic Pan and Scan. Jain {\it et al.}~\cite{jain2015gaze} build upon their work and model the camera trajectories as parametric piece-wise spline curves. However, such parametric motion models have limited applicability due to assumptions on the content (a single pan in each shot). Grundmann {\it et al.}~\cite{grundmann2011auto} show that `professional cameraman like' trajectories consist of piece-wise static, linear, and parabolic segments. They show the applicability of such a motion model for video stabilization. Their approach seems motivated by cinematic ideas, generalizes well, and is deployed on large scale systems like Youtube. A similar formulation has been employed in applications of virtual cinematography in panoramic~\cite{gandhi2014multi} and 360 videos~\cite{tang2019joint}. However, the optimization posed in these works~\cite{grundmann2011auto, gandhi2014multi,tang2019joint} is offline (and non-causal) and in our work, we extend it to online and causal settings.  

Carr {\it et al.}~\cite{carr2013hybrid} proposed a hybrid camera to aesthetic video generation in the context of basketball games. They combined robotic cameras to coarsely track the game and augmented it with virtual camera simulations to get smoother camera movements.  They show that the loss of image resolution can be minimized by using a hybrid system. They investigate a causal moving average filtering and a non-causal $l_1$ trend filtering~\cite{kim2009ell_1} to filter the crude trajectories obtained by following the centroid of player detections. They extend their work by learning per frame pan angle predictors based on the player positions in the game and further smooth it using savitzky Golay filter~\cite{savitzky1964smoothing}. In~\cite{chen2016learning}, they merge the camera position prediction and smoothing into a unified framework. These works~\cite{kim2009ell_1, chen2016learning} rely on a supervised signal generated by a synchronized human camera operator, which is difficult to obtain and also makes them domain-specific. In contrast, the proposed filter in our work is unsupervised. 

Our work is also related to the work of Liu {\it et al.}~\cite{liu2016meshflow}, which proposed a framework for online video stabilization of casually captured videos. Their method performs video stabilization by optimizing vertex level motion trajectories. It runs with a single frame latency and runs optimization locally at every frame using a few previous frames. However, such exact optimization only applies to minimal cost functions like mean squared error + L2 norm smoothness (as in~\cite{liu2016meshflow}). Such minimal functions fail to give the desired cinematic behavior. Moreover, running the optimization on every frame is computationally expensive when a closed-form solution does not exist. We tackle this problem by controlling strides in a sliding window optimization. The proposed \emph{CineConvex} filter: (a) uses a cinematically motivated cost function, and (b) provides more structure due to historical constraints and peek into the future frames.  Moreover, beyond a standalone optimization for each window, the proposed \emph{CineCNN} filter can learn priors/structures from previous data as well as temporal dependencies within a sequence.

Our work, limits to the problem of one-dimensional trajectory filtering/stabilization. The work is agnostic to the application and is directly applicable in variety of frameworks~\cite{yokoi2005virtual, sun2005region, chen2010personalized, heck2007virtual, carr2013hybrid, chen2016learning, liu2016meshflow, gandhi2014multi, zhang2008automated}. We compare our approach against trajectory filtering approaches employed in these frameworks.

\section{The CineFilter Model}
In this section, we describe our models and their unsupervised optimization procedure. Before going into the problem formulation, we list the desiderata that guide our modeling choices. \vspace{-0.5em}
\begin{enumerate}
    \item The smoothing filter should be online and run in real-time. \vspace{-0.5em}
    \item It should not require labeled supervisory data (e.g. ground truth trajectories from human experts).\vspace{-0.5em}
    \item Unmotivated camera movements should be avoided. For instance, when the subject in the frame is stationary, the camera behavior should be static to ensure a pleasant viewing experience.\vspace{-0.5em}
    \item Trajectory smoothing should be robust to outliers in the case of sudden changes in camera trends (should avoid abrupt jerks).\vspace{-0.5em}
    \item It should avoid the accumulation of drift, which is a common problem observed in real-time systems.
\end{enumerate}

With the above noted, we first define our problem as that of predicting a smooth camera trajectory given a stream of incoming noisy trajectory positions. Let $X_t = \{x_0, x_1, ...x_t\}$ be the noisy input sequence that has arrived until frame $t$ and $Y_{t-1} = \{y_0, y_1, ...y_{t-1}\}$ be the smoothed output sequence predicted until the previous frame $t-1$. We wish to learn a nonlinear causal filter specified by a parametric model $y_t = f(X_t, Y_{t-1})$, which predicts the smoothed output for the current frame $t$. At any timestep, predicting the next smooth frame is ideally a function of both the past and future trajectory directions, which is feasible to model using offline approaches. In the online scenario that we wish to operate in, learning the above mapping is an ill-posed problem as multiple solutions can exist depending on where the future trajectory goes. Therefore it makes sense to accommodate a small number of future frames into the model to constrain the direction of the trajectory.

In this work, we propose two different methods to solve the above problem. The first is an online filter that performs convex optimization on a combination of objectives. Since the final objective is convex, we can run a per-sample solver to obtain a global minimum at each timestep. However, these methods have a dependency on the solver at deployment time and are computationally intensive if run at each frame in an online fashion. These methods also do not learn a data-based model of trajectory behavior; hence, they might not be fit for cases with high variance in data statistics. The second model we propose is a learning-based solution combining total variation denoising with a 1D convolution neural network (CNN). The CNN gives advantage over convex optimization approach as it can learn trajectory patterns from data. Moreover, at inference, a forward pass through this model is faster and computationally cheaper than a convex solver.

The ideal camera trajectory should be composed of three types of segments, namely static segments, constant velocity segments, and segments with constant acceleration, all transitioning in a smooth manner. As opposed to previous works that use ground truth data from human operators or create large datasets for deep learning based methods, we build our model through an unsupervised multi-objective loss function that enforces such behavior without the need to collect labeled data.

\subsection{CineConvex filter}
\label{convex_model}
\begin{figure}[t]
    \centering
    \includegraphics[width=0.9\linewidth]{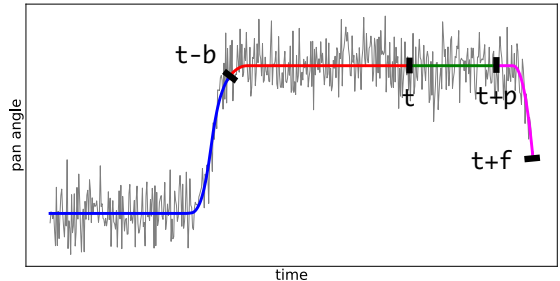}
    \caption{The sliding window configuration for our CineFilter models. At timestep $t$, the model stores $b$ timesteps of the past buffer (in red), has access to $f$ timesteps of the future (in pink) and shifts with a stride of $p$ after each prediction (in green).} 
    \label{fig:window}
\end{figure}


We first describe the convex optimization-based solution for online smoothing. The model works in a sliding window manner. The sliding window configuration is shown in Figure~\ref{fig:window}. It consists of three fragments. The first fragment is called the present window of size $p$ from timestep $t$ to $t+p$ where $t$ is the current time. This fragment gets updated in the final predictions after optimizing the current sliding window. It is also the step-size by which we shift the window after each optimization. The second fragment is called the buffer window, which spans timestep $t-b$ to $t$. It is the historical trajectory information we use in the optimization. The third fragment is called the future window which holds timestep $t$ to $t+f$ with $t+f>t+p>t$ and includes the present window. It provides future context for the optimization. We optimize the trajectories from timestep $t-b$ to $t+f$ and shift all windows by $p$ after each optimization. 

The optimization procedure for each sliding window includes: (a) a term to enforce the predicted trajectory to be close to the original trajectory in some distance metric and (b) L1-norm of the first-order, second-order and third-order derivatives over the optimized trajectory to induce piece-wise static, linear and parabolic behavior. L1-norm has the property to avoid residual motions (e.g., when the path is meant to be static, it leads to truly static outputs) and avoids the superposition between the constant, linear, and parabolic segments.  The final objective $J^m$ with respect to timestep $t$, where $m$ indexes the $m^{th}$ optimization procedure, is given by:

\begin{equation} \small
    \label{eqn:convex_secondtorder}
    J^m(t) = \lambda_0 D_0^m(t) + \lambda_1 D_1^m(t) + \lambda_2 D_2^m(t) + \lambda_3 D_3^m(t)
\end{equation}
where,\\
\begin{equation} \small
    \label{eqn:convex_mseloss}
    D_0^m(t) = \sum_{i=t-b}^{t+f} (x^m(i) - y^m(i))^2
\end{equation}
\begin{equation} \small
    \label{eqn:convex_firstorder}
    D_1^m(t) = \sum_{i=t-b}^{t+f} \mid y^m(i+1) - y^m(i)\mid
\end{equation}
\begin{equation} \small
    \label{eqn:convex_secondtorder}
    D_2^m(t) = \sum_{i=t-b}^{t+f} \mid y^m(i+2) - 2y^m(i+1) + y^m(i) \mid
\end{equation}
\begin{equation} \small
    \label{eqn:convex_thirdorder}
    D_3^m(t) = \sum_{i=t-b}^{t+f} \mid y^m(i+3) - 3y^m(i+2) + 3y^m(i+1) - y^m(i)\mid
\end{equation}
The $\lambda$s are hyperparameters found using cross-validation and $p$, $b$ and $f$ are values we fix heuristically.

Note that this \emph{CineConvex} model has a latency of $f$ timesteps, stores an extra $b$ timesteps from the past, and is run after every $p$ timesteps. Reducing $f$ can reduce the latency, but will lead to less smooth results since multiple future trajectory directions are possible. Decreasing $p$ will lead to frequent trajectory updates but will affect the speed of the filtering operation due to the larger number of optimizations required. Finally, increasing $b$ will provide further historical context; however, that will increase the time required for each individual optimization. Hence the window-related parameters offer a way to balance the speed and accuracy trade-off.

Another problem encountered during this optimization is to maintaining continuity over optimizations on consecutive sliding windows. To mitigate this, we place a hard equality constraint on the past trajectories being currently predicted ($y^m(t-b)$ to $y^m(t)$) and the past trajectories already predicted during the previous optimization ($y^{m-1}(t-b)$ to $y^{m-1}(t)$). Finally, the objective is changed to the following:
\begin{equation} \small
    \label{eqn:convex_final}
    \begin{aligned}
    J^m(t) = \lambda_0 D_0^m(t) + \lambda_1 D_1^m(t) + \lambda_2 D_2^m(t) + \lambda_3 D_3^m(t) \\ \text{s.t., } y^m(k) = y^{m-1}(k) \text{ } \forall k \in \{t-b, ..., t\}
    \end{aligned}
\end{equation}

We use the state of the art Gurobi solver~\cite{solver2019gurobi} for minimizing the objective function (the fastest solver in the MIPLIB 2017 Benchmark~\cite{miplib2017}). 

\subsection{CineCNN filter}
\label{lstm_model}
The \emph{CineConvex} model described above allows us to obtain a global minima for a given fragment, but there are two potential issues with it. Firstly, the optimization is performed on a per-sliding window basis and needs to run frequently during the online operation of the filter. In addition to computational burden and it has a vital dependency on the speed of the solver. Secondly, since there is no data-based learning involved and the optimization described only works at window-level trajectories, the model cannot build a global model for the variation in data statistics that can be encountered by the model. Moreover, since the objective is always an approximation to the behavior we want in the real world, having some inductive biases directly from the data might help in learning better models. This motivates our use of a data-driven model to solve the smoothing problem.

We can pose the problem of estimating the current smooth trajectory position as a sequence modeling problem. However, popular sequence prediction models like HMMs, RNNs, and LSTMs operate sequentially and have a high memory footprint. Moreover, we found in our experiments that 1D encoder-decoder CNN based architecture better learns the local structure over the more complex recurrent counterparts and also provides improved performance. To promote perfectly static trajectories, wherever possible, we first filter the input signal using a 1D Total Variation (TV) denoising algorithm~\cite{condat2013direct} and then pass it to the CNN. The TV output leads into staircase artefacts whenever the camera movement occurs. The CNN smooths out the staircase artefacts and the results into trajectories that mimic professional camera behavior (having smooth transitions between perfectly static segments). We use an extremely fast direct (non iterative) TV denoising algorithm~\cite{condat2013direct}. In the absence of smooth trajectory labels, we use a similar unsupervised loss function for training the CNN as the convex optimization, without the first-order term. The loss function penalizes the squared distance between the original trajectory and the predictions along with the second, and third-order derivative terms.


The final loss function for the model is as follows:
\begin{equation} \small
    \label{eqn:convex_secondtorder}
    L = \lambda_0 D_0  + \lambda_2 D_2 + \lambda_3 D_3
\end{equation}


The first three terms $D_1, D_2, D_3$ are the same as before, but from timestep $1$ to $n$.

While training, the input sequence from all trajectories are divided into overlapping subsets of $n=512$ frames. The inference happens on a sliding window of 32 frames, which is possible since the network is fully convolutional. During inference, when only the first frame has arrived, we left-pad the input with repeated values of the first trajectory position and use this as the input. Since there is no optimization at test time, but only a filter-forward pass through the model, we can make a prediction at each time step i.e., with $p=1$ in figure~\ref{fig:window}. Also, note that there is no explicit constraint that enforces trajectory continuity across predictions like the one needed in the convex optimization formulation. The model has a structural inductive bias (1D convolution) that merges information from local trajectory positions, thus aiding continuity. The data itself, which is fed as a sequence, also provides an additional implicit constraint on the smoothness of the predictions.

\subsection{Implementation details}
For \emph{CineCNN}, we use a 5-layered 1D encoder-decoder based CNN model that has skip connections similar to U-Net architecture with a kernel size of $3$ and with $16$, $32$, $32$, $16$ and $1$ filters respectively. Each CNN layer has a relu based activation on the outputs. The videos from TLP dataset~\cite{moudgil2017long} and Basketball dataset~\cite{chen2016learning} (described in Section~\ref{dataset_description}) are used to train the model. We sub-sample from the full trajectory of the video at random frames and create a set of input trajectories each of a fixed length of $512$ and use these as single instances of training data for the model. 

We create $98k$ such instances through sub-sampling. The model is trained for $20$ epochs on basketball and TLP datasets, each with a batch size of $16$. The adam optimizer is used during training. We follow a learning rate schedule that decays the learning rate by a factor of $0.1$ if the validation loss plateaus for more than $4$ epochs. The weights associated with each loss term are obtained empirically and $(\lambda_0, \lambda_1, \lambda_2, \lambda_3)$ for stage performance dataset are $(1.0, 1000, 50, 2000)$ and for basketball dataset are $(1.0, 2000, 100, 3000)$. The values for $p, f, b$ are $8, 16, 64$, respectively, for the \emph{CineConvex} model. For the \emph{CineCNN}, we set $p=1$, $f=16$ and $b=16$. 

\section{Experiments}
\label{experiments}

We evaluate our approach for the automated broadcast of basketball matches~\cite{chen2016learning} and staged performances~\cite{gandhi2014multi}. We compute the results of \emph{CineConvex} and \emph{CineCNN} with four algorithms commonly used for online filtering in previous works. We also perform ablation studies to demonstrate the effect of the parameters $p$, $f$, $b$ on model performance. We now describe the datasets, baselines, and evaluation strategy in detail.


\subsection{Datasets}
\label{dataset_description}

{\bf Basketball dataset:} We use the Basketball dataset proposed by Chen et al.~\cite{chen2016learning}. This dataset consists of a video recording of a high school basketball match taken from two different cameras. One wide-angle camera is installed near the ceiling and looks at the entire basketball area. The feed from the wide-angle camera is used to detect players and compute features summarizing the current state of the scene. The second broadcast camera is placed at the ground level and is manually operated by a human expert. The evaluation task is to predict the pan angle for a robotic camera, given the current state of the match observed by the wide camera. The pan angle of the human-operated camera is considered as the ground truth (calculated by computing its homography with respect to the wide-angle camera). The dataset consists of 50 segments of 40 seconds each (overall 32 minutes of in-play data), out of which 48 segments are used for training, and 2 are used for validation and testing. Similar to~\cite{chen2016learning}, we train a Random Forest regressor to obtain per frame pan angle predictions. For learning the Random Forest regressor, the game features computed using the wide-angle camera are used as input and the human operator pan angle as ground truth labels. The per-frame predictions give an extremely noisy output, which are then subjected to a filtering operation. We perform comparisons on the output of CineFilter models and the baselines with the human operator trajectory as the ground truth. 
 


{\bf Stage Performance dataset:} We build a Stage Performance dataset that comprises of two wide-angle recordings of staged performances each of $12$ and $10$ minutes, respectively (a dance and a theatre performance). The videos are selected from the Track Long and Prosper (TLP) Dataset~\cite{moudgil2017long}. The original recordings were done using a static wide-angle camera covering the entire action in the scene. The noisy object trajectory sequences for these recordings are obtained using the MDNet tracker~\cite{nam2016learning}. A complete $12$ minute sequence is used for training, and the $10$ minute sequence is used for testing. The filters are evaluated on the task of virtual camerawork, following an actor on the stage based on the output of the tracker. Since there is no ground truth available for these videos, we use the offline optimizer from~\cite{gandhi2014multi} as the ground truth trajectory. 
\subsection{Baselines}

{\bf Savitzky Golay: } SG filter performs data smoothing using least squares to fit a polynomial of a chosen degree within a window of consecutive data points around every point. It takes the central value in the window as the new smoothed data point and this step is performed at each point on the time series. SG filter is non-causal and in our case we choose a window size of $51$, giving a latency of $25$ frames. The degree of the polynomial is set to $3$. 

{\bf Kalman Filter: } Kalman Filter is a recursive Bayesian estimation method that can be used for updating the current smooth camera state at every time step using previous predictions and the current observation. We set the parameters of the filter similar to~\cite{chen2016learning}. 

{\bf Bilateral Filter:} Bilateral filter is a non linear, adaptive, edge preserving and noise reducing smoothing filter. It uses a weighted-average approach where the weights at any given data point is computed using two Gaussians, one Gaussian on the difference of spatial distance and the other on the difference between the magnitudes of the given point with every other point in the surrounding window. It is also non-causal and we use a window size of $64$, giving the latency of $32$ frames. 

{\bf MeshFlow:} MeshFlow~\cite{liu2016meshflow} runs with a single frame
latency and runs optimization locally at every frame using a few previous frames. The optimization functions consists of two different terms. The first term is the MSE over the predicted and original values of the 1D signal. The second term is a MSE over the first order differential of the predicted signal. Window size of 20 values is considered for the experiments.  


\subsection{Evaluation Metric}


For quantitative experiments on the Basketball dataset, we use the data from the human operator as ground truth and compute two different metrics, one measuring the closeness to the original signal and other measuring the movement profile. The first metric (precision metric) is the mean squared error between the filtered trajectory predictions and the ground truth trajectories (pan angle corresponding to the expert human operator). Assuming that the human operator selects the pan angle that best showcases the activity happening in the game, this term penalizes trajectories that deviate from the angles at which important activities are happening. The second metric (smoothness metric) is the absolute difference in the slopes between the predicted and the ground truth trajectory. It measures the ability to mimic human cameraman-like behavior. It penalizes any movement of the predicted trajectory when the human operator is static and also penalizes the predicted trajectory if it moves in a different direction or at a different speed than the ground truth. We define the two terms as Precision Loss and Smoothness Loss:


\begin{equation} 
    \label{eqn:precision}
    precision = \sum_{t}^{}(y(t) - \hat{y(t)})^{2} \\
\end{equation} 

\begin{equation}
    \label{eqn:smoothness}
    smoothness = \sum_{t}^{} \mid \frac{dy(t)}{dt} - \frac{d\hat{y(t)}}{dt} \mid
\end{equation}

$y(t)$ is the prediction, $\hat{y(t)}$ is the ground truth from the human operator, lower the precision \& smoothness loss, the better the predictions.

Although quantitative metrics can give a reasonable estimate about the effectiveness of the predictions, the final gold standard is the human perception of the rendered videos using the filtered trajectories. For instance, an aesthetically pleasing viewing experience is more important even if it comes at the cost of increased precision loss. Hence, in addition to quantitative metrics, we also evaluate our model qualitatively in the form of a comprehensive user study. The study is done on the output shots obtained from the task of virtual camera simulation on 14 small sequences from different videos of stage performances. 


\section{Results}
\label{results}

\subsection{Quantitative Evaluation}
\label{quant}
\begin{figure}[t]
    \centering

    \includegraphics[width=0.49\linewidth]{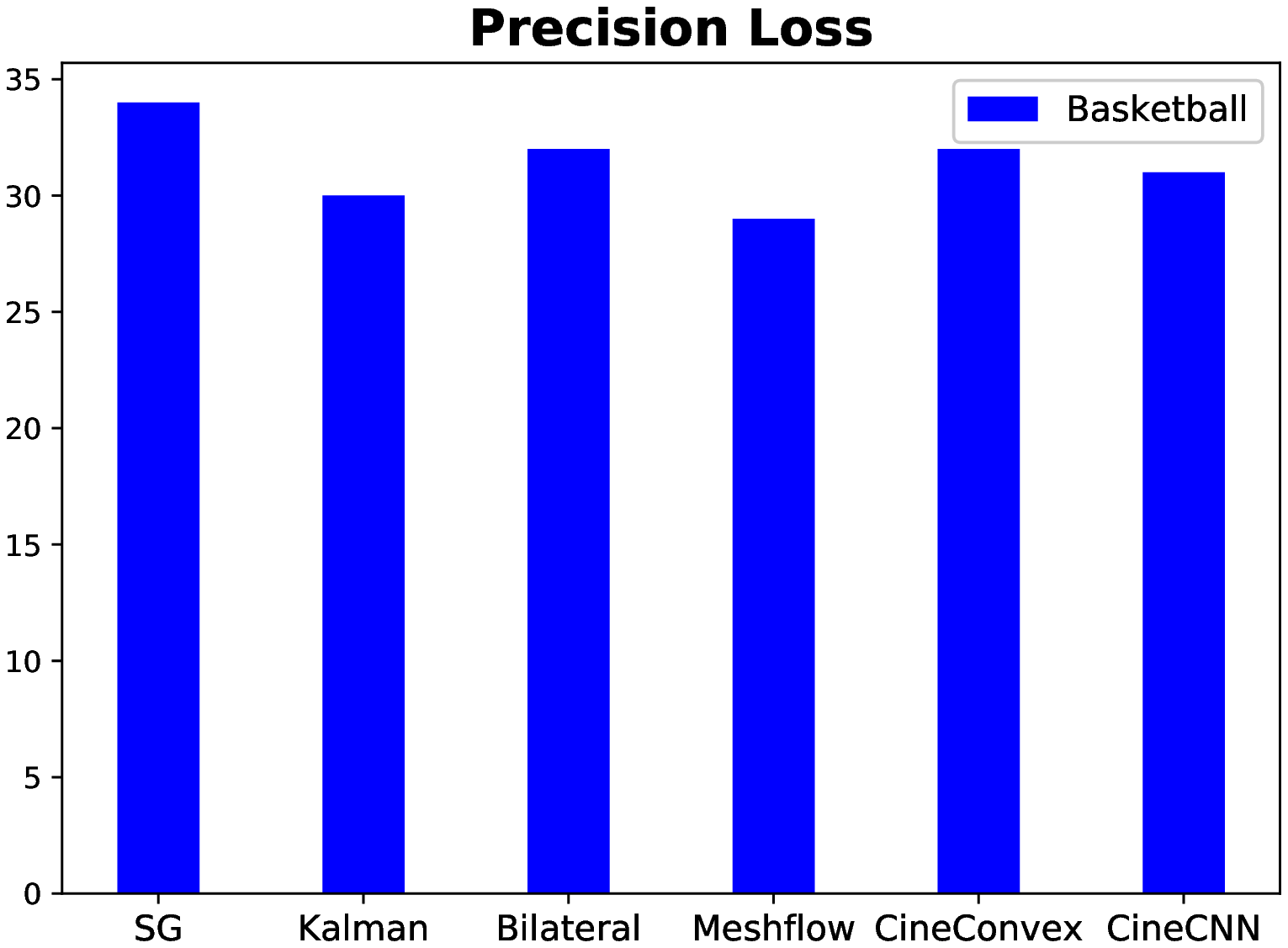}\hspace{0.03em}\vspace{0.1em}
    \includegraphics[width=0.49\linewidth]{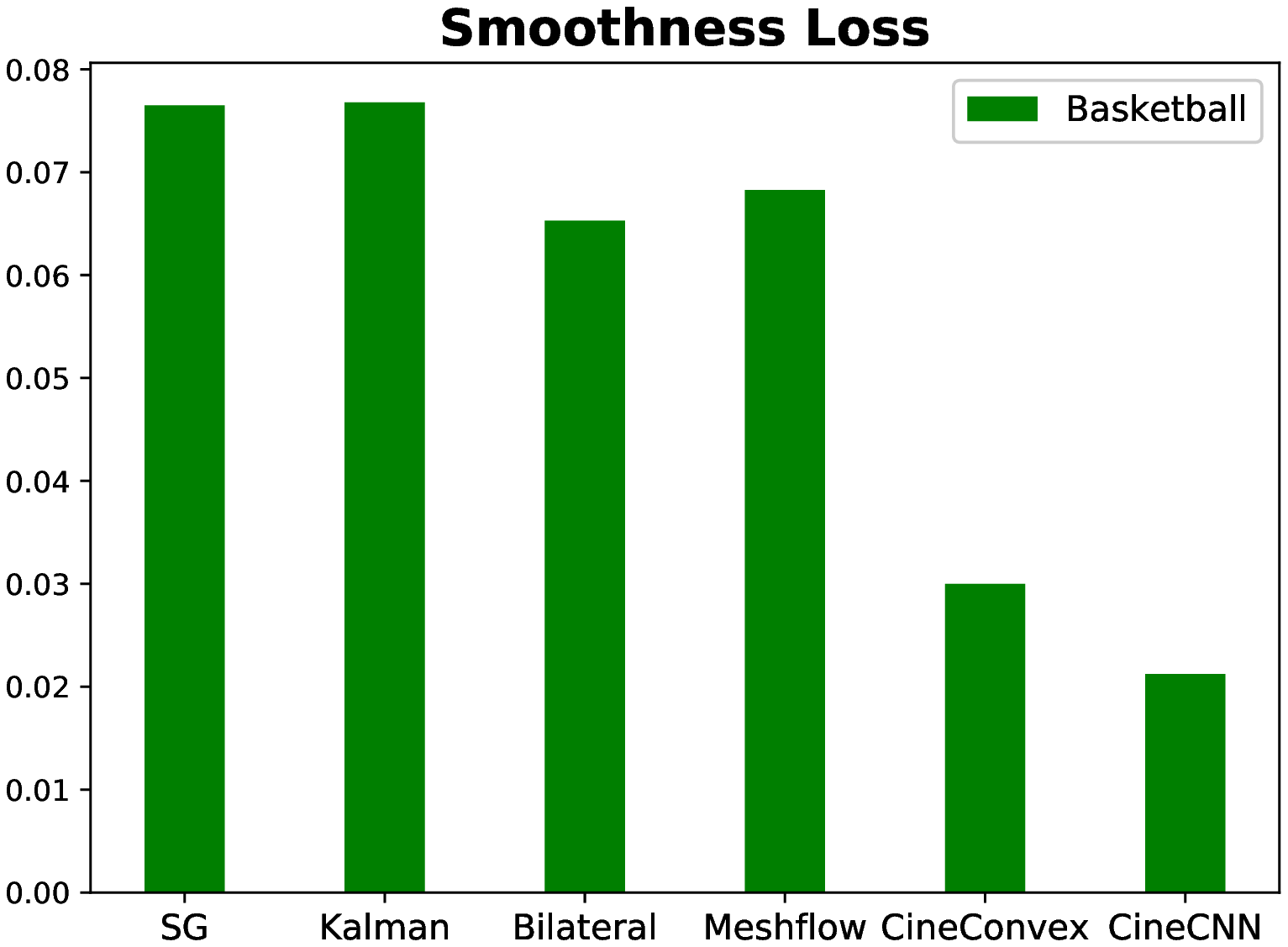}\vspace{0.1em} \\
    \includegraphics[width=0.49\linewidth]{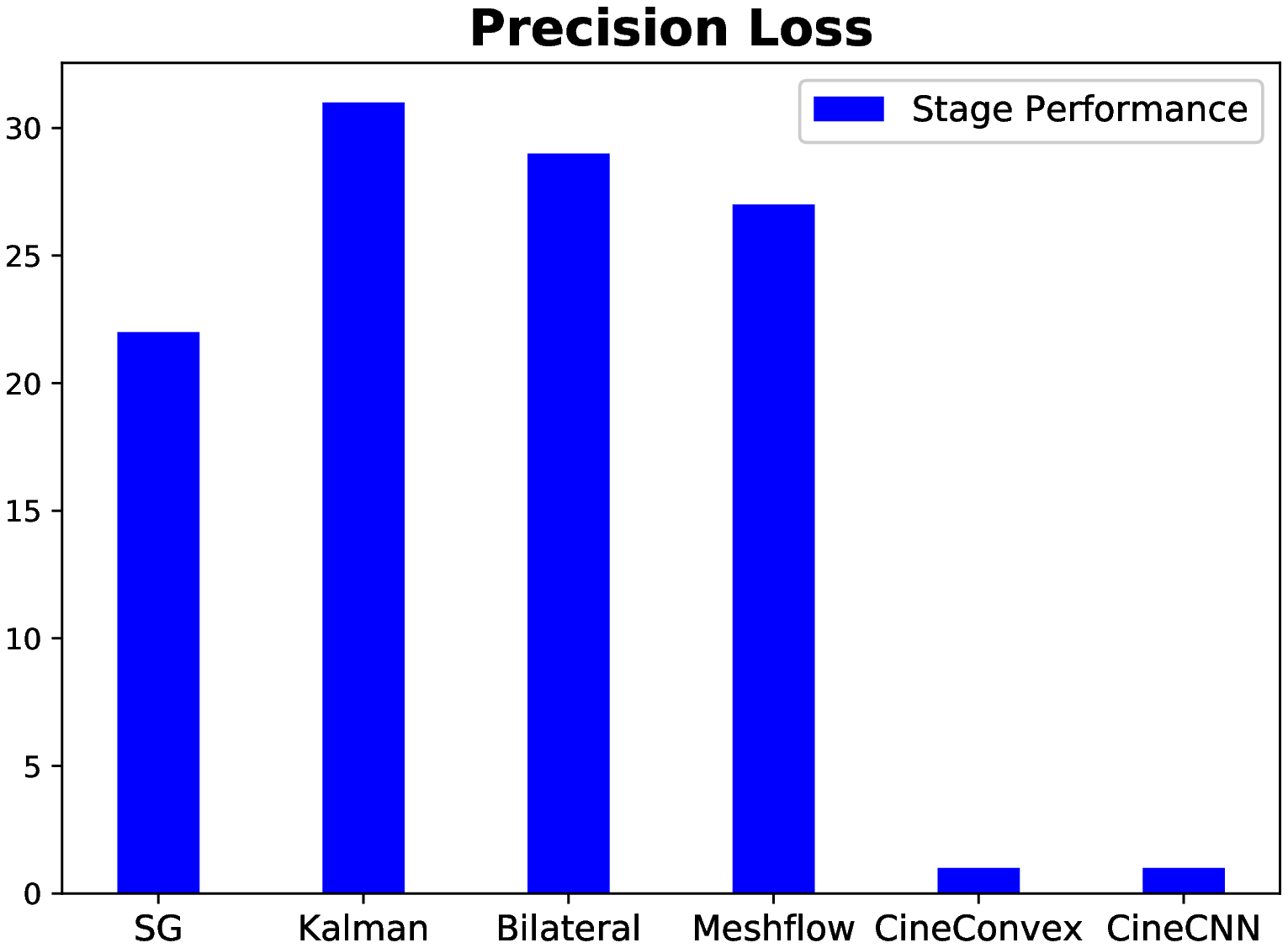}\hspace{0.03em}
    \includegraphics[width=0.49\linewidth]{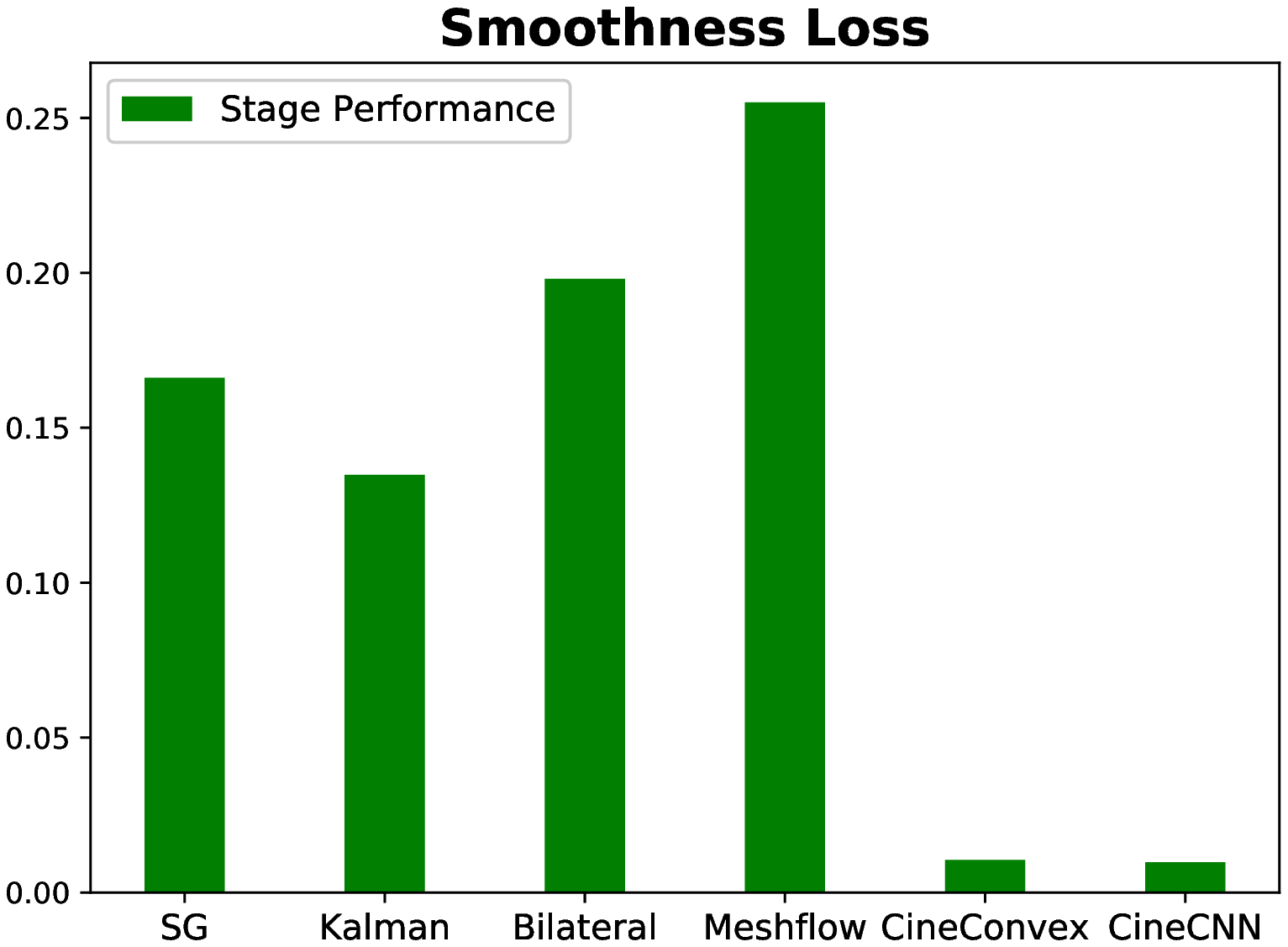} 

    \caption{Precision \& Smoothness loss for our approach and the baselines on the Basketball dataset (top row) and the Stage Performance dataset (bottom row).} 
    \label{fig:quantitative_result_stage}
\end{figure}

We compare \emph{CineConvex} and \emph{CineCNN} against the baselines on the precision loss (Equation~\ref{eqn:precision}) and smoothness loss (Equation ~\ref{eqn:smoothness}) metrics. The results are summarized in Figure~\ref{fig:quantitative_result_stage}.

For the Basketball dataset, which has the noisier motion of the two datasets, we see that the proposed models are competent on the precision metric to other approaches; however, they bring significant improvements over the smoothness metric. \emph{CineCNN} gives more than three times improvement over the other baselines. We also observe that \emph{CineCNN} gives better performance in high noise situations over the \emph{CineConvex} filter. 

For the Stage Performance dataset, which has relatively noise-free trajectories, \emph{CineConvex} and \emph{CineCNN} give similar performances. The proposed methods notably outperform the baselines over both precision and smoothness metric. The use of a ground truth generated using similar loss terms (offline) might give an added advantage to the proposed models. However, offline optimization~\cite{grundmann2011auto, gandhi2014multi} is shown to extremely effective and considered to closely mimic human cameraman behaviour. The efficacy of the proposed models (over baselines) is further affirmed by the user study presented in the following section.




\subsection{User Study and Visual Inspection}
\label{qual}
Although the precision \& smoothness loss is a reasonable way to assess our model, it may not be able to measure the adherence to cinematic principles accurately and, more importantly, the final aesthetics of the rendered video. For instance, a smoother trajectory may be preferred by the user even if it is slightly drifted and increases the precision loss. Similarly, unmotivated movement can appear distracting to the user, even if they are extremely minute and may not significantly contribute to the loss. To this end, we complement our evaluation using qualitative methods to account for the perceptual metrics i.e., how the proposed filtering method performs against the baselines in terms of aesthetics of the rendered video; we perform a study with $14$ users. 

We select $14$ small video clips from a diverse set of wide-angle stage recordings, which are different from the two sequences in Stage Performance dataset (used in training). The average duration of the clips is 25 seconds. We evaluate each of the proposed and baseline filters for the virtual camera simulation task~\cite{gandhi2014multi}. The filters are applied over the per frame shot estimations obtained from noisy actor tracks. 

In each trial, the participants are shown two videos in a side by side manner, one rendered using \emph{CineCNN} and the other using \emph{CineConvex} or one of the baselines. They are instructed to choose the video that is more aesthetically appealing and better mimics human camerawork. They are also given an option to choose neither (if they do not have a clear preference, and both videos are reasonably similar in terms of aesthetics). Each user watches nine pairs of videos; therefore, each pair of videos is watched exactly once. The left and right ordering for videos is randomly switched. The results are illustrated in Table~\ref{tab:user_study_cnn}.

\begin{table}
    \centering
    \begin{tabular}{l|*{4}c}
        \toprule
        CNN & Win & Loss & No Preference\\
        \midrule
        vs Kalman & 13 & 0 & 1 \\ 
        vs SG & 14 & 0 & 1 \\
        vs Meshflow & 11 & 1 & 2 \\
        vs Bilateral & 11 & 1 & 2 \\
        vs Gurobi & 3 & 0 & 11 \\
        \bottomrule
    \end{tabular}
\caption{User study results of baselines approaches vs \emph{CineCNN} filter.}
\label{tab:user_study_cnn}
\end{table}

\begin{table}
    \centering
    \begin{tabular}{l|*{4}c}
        \toprule
        Gurobi & Win & Loss & No Preference\\
        \midrule
        vs Kalman & 12 & 1 & 1 \\ 
        vs SG & 14 & 0 & 1\\
        vs Meshflow & 11 & 1 & 2 \\
        vs Bilateral & 10 & 1 & 3 \\
        vs CNN & 0 & 3 & 11 \\
        \bottomrule
    \end{tabular}
    \caption{User study results of baselines approaches vs our \emph{CineConvex} filter.}
    \label{tab:user_study_cvx}
\end{table}

The proposed methods are significantly preferred compared to the other baselines. Bilateral is the most competent approach among the baselines, and it works well in sequences where actors are continually moving. Kalman filter has minimal drift and is preferred in a few cases, as it maintains the shot compositions well. We present some of these rendered comparison videos in the supplementary material. \emph{CineCNN} is consistently chosen over \emph{CineConvex}, which correlates with the precision \& smoothness loss shown in Figure~\ref{fig:quantitative_result_stage}. On the other hand, since the Basketball dataset does not have publicly available video sequences, we point the readers to the predicted trajectory comparison for all the baselines and between \emph{CineConvex} and \emph{CineCNN} models in Figure~\ref{fig:teaser}. The proposed \emph{CineCNN} filter outperforms other methods on the Basketball dataset (in terms of smoothness, lack of sharp jerks, and lack of residual motion), as also indicated by the precision \& smoothness Score in Figure~\ref{fig:quantitative_result_stage}.

\subsection{Ablative Experiments}
\label{ablative}


In this section, we discuss results for ablation experiments across different values for the window hyperparameters $p$ and $f$ and show performance and speed varies across the two proposed model formulations. Table~\ref{tab:ablation_table_cvx} shows the influence of the present window size ($p$) and future window size ($f$) on the (precision loss, smoothness loss) and speed of \emph{CineConvex} filter. The tables show how increasing the present window width can improve the speed of the filtering operation, but also needs an increase in the future frames, thus increasing latency. Also, there exists a middle ground for the $p$ and $f$ values, which balances the performance with speed. We can contrast this with \emph{CineCNN}, which has a constant speed of around 1000 FPS with $p=1$. For \emph{CineCNN}, the values of $f=(4,8,16,32)$ gets respective precision \& smoothness loss of $((31.1, 0.05), (31.9, 0.03), (31.9, 0.02), (31.6, 0.02))$. Like the \emph{CineConvex}, the \emph{CineCNN} has only very slight improvement for $f=32$ over $f=16$, so we use $f=16$ in our experiments. The ablation experiments again show the various speed-accuracy-latency trade-offs associated with both models across the choice of hyperparameters.

\begin{table}
    \begin{adjustbox}{width=\columnwidth,center}
        \centering
        \begin{tabular}{cc}
            
            \begin{tabular}{l|*{4}r}
                \toprule
                \diagbox{p}{f} & 4 & 8 & 16 & 32 \\
                \midrule
                4 & (51.7, 0.05) & (37.2, 0.05) & (31.7, 0.04) & (32.0, 0.03) \\
                8 & - & (35.9, 0.04) & (31.1, 0.04) & (31.6, 0.03) \\
                16 & - & - & (31.7, 0.04) & (30.1, 0.04) \\
                32 & - & - & - & (31.1, 0.04) \\
                \bottomrule
            \end{tabular} &   
            
            \begin{tabular}{l|*{4}r}
                \toprule
                \diagbox{p}{f} & 4 & 8 & 16 & 32 \\
                \midrule
                4 & 147 & 135 & 125 & 98 \\
                8 & - & 250 & 227 & 172 \\
                16 & - & - & 417 & 333 \\
                32 & - & - & - & 714 \\
                \bottomrule
            \end{tabular}
            
        \end{tabular}
    \end{adjustbox}
    \caption{Ablation study of \emph{CineConvex}, comparing model performance (Precision loss, Smoothness loss) on the left and model speed (frames per second) on the right, across various present window size (p) and future window size (f) combinations. }
    \label{tab:ablation_table_cvx}
\end{table}

\subsection{Pre-processed Evaluation}

The proposed \emph{CineCNN} model performs a TV denoising prior to sending the data through the CNN. For a fair comparison, we pre-process the tracks with TV denoising on other baselines as well and compare the outputs. The results are presented in Table~\ref{tab:preprocessed_table}. We observe that TV denoising brings minor improvements in precision and smoothness metrics for the baselines, however, significantly behind the performance of CineCNN. The baselines filters also fail to tackle the staircase artefacts and lead to jerks. 


\begin{table}
        \centering
            \begin{tabular}{l|*{3}c}
                \toprule
                Baselines & Precision & Smoothness \\
                \midrule
                Kalman & 29.1 & 0.144 \\ 
                SG &  19.1 & 0.155 \\
                Meshflow & 27.0 & 0.158 \\
                Bilateral & 37.0 & 0.138 \\
                CineCNN & 0.94 & 0.009 \\
                \bottomrule
            \end{tabular}
\caption{Total Variation pre-processed evaluation of \emph{CineCNN} vs baseline approaches on the Stage Performance dataset.}
\label{tab:preprocessed_table}
\end{table}

\subsection{Residual Motion }

According to professional cinematographic practices~\cite{thompson2013grammar}, a steady camera behaviour is necessary for pleasant viewing experience. A camera movement without enough motivation may appear irritating to the viewer, hence
the camera should remain static in case of small and unmotivated movements. Small residual motions (even minuscule) are displeasing. Since baselines filters are not cinematically motivated, they exhibit residual motion. We quantatively show this by computing the smoothness metrics only on parts where the ground truth is perfectly static for at least 4 seconds (i.e 128 frames). The results presented in Table~\ref{tab:residual_loss_table} clearly demonstrate the efficacy of \emph{CineConvex} and \emph{CineCNN} in terms of providing perfectly static camera trajectories.


\begin{table}
    \centering
    \begin{tabular}{l|*{2}c}
        \toprule
        Baselines & Residual Error \\
        \midrule
        Kalman & 0.0609 \\ 
        SG &  0.1509 \\
        Meshflow & 0.0798 \\
        Bilateral & 0.0592 \\
        CineConvex & 0.0005 \\
        CineCNN & 0.0001 \\
        \bottomrule
    \end{tabular}
\caption{Residual motion loss of baseline approaches vs \emph{CineFilter} models on Stage Performance Dataset. }
\label{tab:residual_loss_table}
\end{table}

\section{Conclusion}
In this paper, we propose two unsupervised methods for the online filtering of noisy trajectories. \emph{CineConvex} formulation poses filtering as a convex optimization over individual sliding windows and is solved using iterative convex solvers. \emph{CineCNN} formulation models filtering as a prediction using a convolutional neural network. The unsupervised nature of the proposed methods, high operational frame rates, and low memory footprint make them a generic choice across a large variety of applications. We contrast the performance of the proposed methods in comparison to commonly used online filters, across two diverse applications of basketball games (fast-paced, multiple players) and theatre plays (following individual actors on a stage). Thorough quantitative and qualitative experiments show that the proposed methods give superior performance over the existing filtering techniques and provide a preferable alternative for trajectory filtering in online automated camera systems. 

{\small
\bibliographystyle{ieee}
\bibliography{egpaper_for_review}

\begin{thebibliography}{10}\itemsep=-1pt

\bibitem{ariki2006automatic}
Y.~Ariki, S.~Kubota, and M.~Kumano.
\newblock Automatic production system of soccer sports video by digital camera
  work based on situation recognition.
\newblock In {\em Eighth IEEE International Symposium on Multimedia (ISM'06)},
  pages 851--860. IEEE, 2006.

\bibitem{bianchi2004automatic}
M.~Bianchi.
\newblock Automatic video production of lectures using an intelligent and aware
  environment.
\newblock In {\em Proceedings of the 3rd international conference on Mobile and
  ubiquitous multimedia}, pages 117--123. ACM, 2004.

\bibitem{cantine1995shot}
J.~Cantine et~al.
\newblock {\em Shot by shot: A practical guide to filmmaking}.
\newblock ERIC, 1995.

\bibitem{carr2013hybrid}
P.~Carr, M.~Mistry, and I.~Matthews.
\newblock Hybrid robotic/virtual pan-tilt-zom cameras for autonomous event
  recording.
\newblock In {\em Proceedings of the 21st ACM international conference on
  Multimedia}, pages 193--202. ACM, 2013.

\bibitem{chen2010personalized}
F.~Chen and C.~De~Vleeschouwer.
\newblock Personalized production of basketball videos from multi-sensored data
  under limited display resolution.
\newblock {\em Computer Vision and Image Understanding}, 114(6):667--680, 2010.

\bibitem{chen2016learning}
J.~Chen, H.~M. Le, P.~Carr, Y.~Yue, and J.~J. Little.
\newblock Learning online smooth predictors for realtime camera planning using
  recurrent decision trees.
\newblock In {\em Proceedings of the IEEE Conference on Computer Vision and
  Pattern Recognition}, pages 4688--4696, 2016.

\bibitem{condat2013direct}
L.~Condat.
\newblock A direct algorithm for 1-d total variation denoising.
\newblock {\em IEEE Signal Processing Letters}, 20(11):1054--1057, 2013.

\bibitem{daigo2004automatic}
S.~Daigo and S.~Ozawa.
\newblock Automatic pan control system for broadcasting ball games based on
  audience's face direction.
\newblock In {\em Proceedings of the 12th annual ACM international conference
  on Multimedia}, pages 444--447. ACM, 2004.

\bibitem{gandhi2014multi}
V.~Gandhi, R.~Ronfard, and M.~Gleicher.
\newblock Multi-clip video editing from a single viewpoint.
\newblock In {\em Proceedings of the 11th European Conference on Visual Media
  Production}, page~9. ACM, 2014.

\bibitem{grundmann2011auto}
M.~Grundmann, V.~Kwatra, and I.~Essa.
\newblock Auto-directed video stabilization with robust l1 optimal camera
  paths.
\newblock In {\em CVPR 2011}, pages 225--232. IEEE, 2011.

\bibitem{heck2007virtual}
R.~Heck, M.~Wallick, and M.~Gleicher.
\newblock Virtual videography.
\newblock {\em ACM Transactions on Multimedia Computing, Communications, and
  Applications (TOMM)}, 3(1):4, 2007.

\bibitem{jain2015gaze}
E.~Jain, Y.~Sheikh, A.~Shamir, and J.~Hodgins.
\newblock Gaze-driven video re-editing.
\newblock {\em ACM Transactions on Graphics (TOG)}, 34(2):21, 2015.

\bibitem{kim2009ell_1}
S.-J. Kim, K.~Koh, S.~Boyd, and D.~Gorinevsky.
\newblock $\ell_1$ trend filtering.
\newblock {\em SIAM review}, 51(2):339--360, 2009.

\bibitem{liu2006video}
F.~Liu and M.~Gleicher.
\newblock Video retargeting: automating pan and scan.
\newblock In {\em Proceedings of the 14th ACM international conference on
  Multimedia}, pages 241--250. ACM, 2006.

\bibitem{liu2016meshflow}
S.~Liu, P.~Tan, L.~Yuan, J.~Sun, and B.~Zeng.
\newblock Meshflow: Minimum latency online video stabilization.
\newblock In {\em European Conference on Computer Vision}, pages 800--815.
  Springer, 2016.

\bibitem{miplib2017}
{MIPLIB} 2017, 2018.
\newblock http://miplib.zib.de.

\bibitem{moudgil2017long}
A.~Moudgil and V.~Gandhi.
\newblock Long-term visual object tracking benchmark.
\newblock {\em arXiv preprint arXiv:1712.01358}, 2017.

\bibitem{nam2016learning}
H.~Nam and B.~Han.
\newblock Learning multi-domain convolutional neural networks for visual
  tracking.
\newblock In {\em Proceedings of the IEEE conference on computer vision and
  pattern recognition}, pages 4293--4302, 2016.

\bibitem{pinhanez1995intelligent}
C.~S. Pinhanez.
\newblock {\em Intelligent studios: Using computer vision to control TV
  cameras}.
\newblock 1995.

\bibitem{rachavarapu2018watch}
K.~K. Rachavarapu, M.~Kumar, V.~Gandhi, and R.~Subramanian.
\newblock Watch to edit: Video retargeting using gaze.
\newblock In {\em Computer Graphics Forum}, volume~37, pages 205--215. Wiley
  Online Library, 2018.

\bibitem{savitzky1964smoothing}
A.~Savitzky and M.~J. Golay.
\newblock Smoothing and differentiation of data by simplified least squares
  procedures.
\newblock {\em Analytical chemistry}, 36(8):1627--1639, 1964.

\bibitem{solver2019gurobi}
G.~Solver.
\newblock Gurobi optimization, 2019.

\bibitem{sun2005region}
X.~Sun, J.~Foote, D.~Kimber, and B.~Manjunath.
\newblock Region of interest extraction and virtual camera control based on
  panoramic video capturing.
\newblock {\em IEEE Transactions on Multimedia}, 7(5):981--990, 2005.

\bibitem{tang2019joint}
C.~Tang, O.~Wang, F.~Liu, and P.~Tan.
\newblock Joint stabilization and direction of 360$\backslash$deg videos.
\newblock {\em ACM Transactions on Graphics (TOG)}, 2018.

\bibitem{thompson2013grammar}
R.~Thompson and C.~Bowen.
\newblock {\em Grammar of the Shot}.
\newblock Taylor \& Francis, 2013.

\bibitem{yokoi2005virtual}
T.~Yokoi and H.~Fujiyoshi.
\newblock Virtual camerawork for generating lecture video from high resolution
  images.
\newblock In {\em 2005 IEEE International Conference on Multimedia and Expo},
  pages 4--pp. IEEE, 2005.

\bibitem{zhang2008automated}
C.~Zhang, Y.~Rui, J.~Crawford, and L.-W. He.
\newblock An automated end-to-end lecture capture and broadcasting system.
\newblock {\em ACM Transactions on Multimedia Computing, Communications, and
  Applications (TOMM)}, 4(1):6, 2008.

\end{thebibliography}
}

\end{document}